\documentclass[letterpaper, 10 pt, conference]{ieeeconf}  
\IEEEoverridecommandlockouts                              
\usepackage{amsmath,amsfonts,amssymb,amsthm}
\usepackage[table]{xcolor}
\usepackage{prettyref}
\usepackage{mathrsfs}
\usepackage{graphicx}
\usepackage{wrapfig}
\usepackage{arcs}
\usepackage{subfig}
\usepackage{makecell}
\usepackage{multirow}
\usepackage{booktabs}
\usepackage{algorithm}
\usepackage{diagbox}
\usepackage
[backend=bibtex,
bibstyle=ieee,
citestyle=numeric,
mincitenames=1,
maxcitenames=2,
natbib=true,
doi=false,
isbn=false,
url=false,
eprint=false]{biblatex}
\usepackage[noend]{algpseudocode}
\usepackage[colorlinks,allcolors=gray,hypertexnames=true]{hyperref} 

\algnewcommand{\LineComment}[1]{\State \(\triangleright\) #1}
\algdef{SE}[DOWHILE]{Do}{doWhile}{\algorithmicdo}[1]{\algorithmicwhile\ #1}
\newcommand*{\colorboxed}{}
\def\colorboxed#1#{%
  \colorboxedAux{#1}%
}
\newcommand*{\colorboxedAux}[3]{%
  \begingroup
    \colorlet{cb@saved}{.}%
    \color#1{#2}%
    \boxed{%
      \color{cb@saved}%
      #3%
    }%
  \endgroup
}
\newtheorem{defi}{Definition}

\newtheorem{thm}{Theorem}
\newtheorem{lem}{Lemma}
\newtheorem{rem}{Remark}
\usepackage{mathtools}

\DeclarePairedDelimiter\floor{\lfloor}{\rfloor}

\newrefformat{fig}{Figure~\ref{#1}}
\newrefformat{par}{Section~\ref{#1}}
\newrefformat{appen}{Appendix~\ref{#1}}
\newrefformat{sec}{Section~\ref{#1}}
\newrefformat{sub}{Section~\ref{#1}}
\newrefformat{table}{Table~\ref{#1}}
\newrefformat{alg}{Algorithm~\ref{#1}}
\newrefformat{def}{Definition~\ref{#1}}
\newrefformat{col}{Corollary~\ref{#1}}
\newrefformat{thm}{Theorem~\ref{#1}}
\newrefformat{step}{Step~\ref{#1}}
\newrefformat{ln}{Line~\ref{#1}}
\newrefformat{eq}{Equation~\ref{#1}}
\newrefformat{pb}{Problem~\ref{#1}}
\newrefformat{it}{Item~\ref{#1}}
\newrefformat{te}{Term~\ref{#1}}
\def\Eqref Eq:#1:{\eqref{eq:#1}}
\newrefformat{Eq}{Equation~\Eqref#1:}

\newcommand{\E}[1]{\mathbf{#1}}
\newcommand{\TE}[1]{\textbf{#1}}

\newcommand{\argmin}[1]{\underset{#1}{\text{argmin}}\;}

\newcommand{\WORK}{\mathcal{W}}
\newcommand{\RR}{\mathbb{R}}

\newcommand{\sta}{\E{s}}
\newcommand{\tar}{\E{t}}

\newcommand{\CIRCLE}{\mathcal{C}}
\newcommand{\SWAPGRAPH}{\mathcal{G}}

\newcommand{\vertex}{\E{v}}
\newcommand{\VERTEX}{\mathcal{V}}
\newcommand{\VERTEXL}{\VERTEX_\mathcal{L}}
\newcommand{\EDGE}{\mathcal{E}}
\newcommand{\EDGEL}{\EDGE_\mathcal{L}}
\newcommand{\EDGEI}{\EDGE_\mathcal{I}}
\newcommand{\vacant}{\vertex_\emptyset}

\definecolor{Gray}{gray}{0.85}

\newcommand{\revised}[1]{\textcolor{black}{#1}}
\makeatletter
\newcommand\fs@ruled@notop{\def\@fs@cfont{\bfseries}\let\@fs@capt\floatc@ruled
  \def\@fs@pre{}%
  \def\@fs@post{\kern2pt\hrule\relax}%
  \def\@fs@mid{\kern2pt\hrule\kern2pt}%
  \let\@fs@iftopcapt\iftrue}
\renewcommand\fst@algorithm{\fs@ruled@notop}
\makeatother

\newif\ifsupp
\suppfalse
\title{\Large\bf Multi-Robot Path Planning Using Medial-Axis-Based Pebble-Graph Embedding   \vspace{-10px}}
\author{Liang He$^{1\dagger}$, Zherong Pan$^3$, Kiril Solovey$^2$, Biao Jia$^4$, and Dinesh Manocha$^4$ 
\vspace{-60px}
\thanks{$^1$Liang He is with Department of Computer Science, University of North Carolina at Chapel Hill. \{lianghe.hust@gmail.com\} $^2$Kiril Solovey is with the Technion, Israel Institute of Technology. \{kirilsol@technion.ac.il\} $^3$Zherong Pan is with Lightspeed \& Quantum Studio, Tencent America. \{zrpan@tencent.com\} $^4$Biao Jia and Dinesh Manocha are with Department of Computer Science and Electrical \& Computer Engineering, University of Maryland at College Park. \{biao,dm@cs.umd.edu\}}}
        
\allowdisplaybreaks
\begin{document}
\maketitle
\thispagestyle{empty}
\pagestyle{empty}

\begin{abstract}
We present a centralized algorithm for labeled, disk-shaped Multi-Robot Path Planning (MPP) in a continuous planar workspace with polygonal boundaries. Our method automatically transform the continuous problem into a discrete, graph-based variant termed the pebble motion problem, which can be solved efficiently. To construct the underlying pebble graph, we identify inscribed circles in the workspace via a medial axis transform and organize robots into layers within each inscribed circle. We show that our layered pebble-graph enables collision-free motions, allowing all graph-restricted MPP instances to be feasible. MPP instances with continuous start and goal positions can then be solved via local navigations that route robots from and to graph vertices. We tested our method on several environments with high robot-packing densities (up to $61.6\%$ of the workspace). For environments with narrow passages, such density violates the well-separated assumptions made by state-of-the-art MPP planners, while our method achieves an average success rate of $83\%$.
\end{abstract}
\section{Introduction}
We propose a centralized approach for Multi-Robot Path Planning (MPP) for planar workspaces with polygonal obstacles. This problem has a wide range of applications such as warehouse management \cite{ma2017lifelong}, computer games \cite{samvelyan2019starcraft}, and crowd modeling \cite{malinowski2017multi}, which require the coordination of a large swarm of robots within a limited computational budget. MPP plans must be computed within a couple of milliseconds for an interactive game and an automatic warehouse needs to answer thousands of queries on a daily basis. Unfortunately, solving general MPP problems is NP-hard \cite{doi:10.1177/027836498400300405,10.1016/j.tcs.2005.05.008,SPIRAKIS198455} and practical algorithms based on sampled roadmaps \cite{Le_Plaku_2017} and conflict-based search \cite{sharon2015conflict} quickly become intractable for more than a few robots. Follow-up research relies on additional assumptions on environment shapes and/or robot arrangements to attain practical performance. 

\subsection{Related Work}
We review the three assumptions most relevant to our work. To begin with, graph pebbling \cite{auletta1999linear,MOEWS1992244,yu2015pebble} lays the theoretical foundation of discrete structure in MPP problems. It assumes that robots are restricted to vertices and move along the edges of a graph. Prior work established fast algorithms to verify and construct feasible solutions for sub-classes of pebble-graphs \cite{auletta1999linear,yu2015pebble}. Certain regular grids are endowed with complete results \cite{standley2011complete,sharon2012meta,sharon2015conflict,6630671} and near-optimal solutions \cite{Yu19AR,yu2018effective} although exact optimality is intractable to attain \cite{Yu2015OptimalMP}. However, building a pebble-graphs for workspaces with complex boundaries is still challenging, because regular grids cannot cover narrow passages and sharp features as illustrated in \prettyref{fig:pipeline} (b).

Some methods \cite{solovey2014k,krontiris2014rearranging,otte2018dynamic,atias2018effective,dayan2020near,shome2020drrt} use random sampling to build discrete graphs that capture the structure of the continuous problem.  
Some of these methods have completeness guarantees, that is, if a graph leading to feasible solutions exists, they can ultimately build one. However, to the best of our knowledge, those approaches do not scale well with the number of robots. 

Lastly, the well-separated assumption \cite{adler2015efficient,solomon2018motion,SolYuZamHal15RSS,tangcomplete} partially addresses the shortcomings of the above methods. By assuming robots (as well as their goal positions) are sufficiently separated from each other and the obstacles, each robot can find paths to their goal regardless of the order of movements of other robots. Unlike graph-pebbling, the well-separated assumption can be verified and established for an arbitrary workspace easily, after which feasibility can be guaranteed. On the downside, the large separation distance can limit the number of robots.

\subsection{Main Results}
\begin{figure*}[th]
\centering
\includegraphics[width=0.96\linewidth]{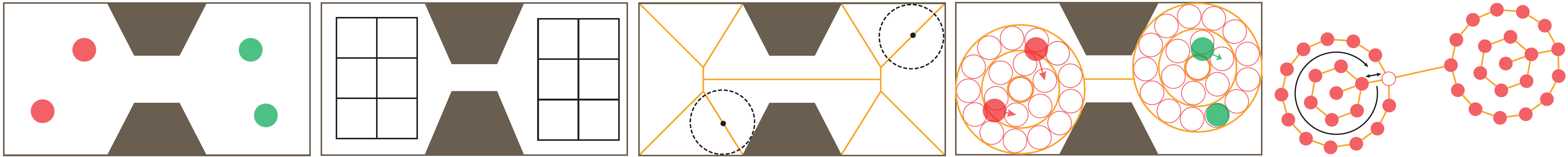}
\put(-465,9){\small{(a)}}
\put(-368,9){\small{(b)}}
\put(-290,20){\small{(c)}}
\put(-170,9){\small{(d)}}
\put(-50,9){\small{(e)}}
\caption{\label{fig:pipeline} \revised{(a): We have two robots moving from start (red) to target (green) positions. (b): Prior work such as \cite{YuLav16TOR} assumes that robots can move on a regular grid (black edges). (c):To extract the
skeletons (yellow), our method first computes a medial axis transform, each point on which is the center of an inscribed circle. (d): Our method embeds the pebble-graph into $\WORK$ by picking two inscribed circles, in which robots are arranged into loops (separated by yellow loop boundaries). We further use subsets of the skeleton as paths connecting inscribed circles. Generally speaking, $\sta^i$ and $\tar^i$  do not coincide with the planned positions in the inscribed circles and we use local navigation to move robots there (arrows). (e): The topology of the embedded graph consists of six connected loops, on which a robot can move to a neighboring vacant position and robots on a loop can perform rotational motions (black edges).}}
\vspace{-15px}
\end{figure*}

We propose a new method to construct pebble-graphs for arbitrary workspaces without using sampling-based methods. As illustrated in \prettyref{fig:pipeline} (c), our method relies on the medial-axis transform \cite{giesen2012medial} to identify a series of inscribed circles with center points connected into skeleton curves. We then organize robots into discrete layers inside each inscribed circles, as illustrated in \prettyref{fig:pipeline} (d). We show that our robot arrangement allows both translational and rotational pebble motions. Following an idea similar to \cite{yu2015pebble}, we prove that all the MPP instances restricted to our pebble-graph are feasible as long as the number of graph vertices is larger than the number of robots. Finally, our method can inherently extend to continuous start and goal positions by moving robots to and from closest graph vertices via local navigation algorithms \cite{van2011reciprocal}.

Our method works well in workspaces with complex obstacles and boundary shapes. We have conducted systematic comparisons with \cite{yu2018effective} in $5$ different environments, each having $30-160$ randomized MPP instances. For the benchmark with narrow passages, our method achieves a $100\%$ success rate on an average robot density between $30-71\%$ of the workspace, while such a high density violates the well-separated assumption in \cite{solomon2018motion} and the method in \cite{yu2018effective} can fail whenever agents fall inside these narrow passages.

\section{\label{sec:problem}Problem Statement and Approach Overview}
\revised{We formalize 2D labeled MPP problems. We define $\WORK\subseteq\RR^2$ as the 2D workspace and assum that the boundary of workspace, $\partial\WORK$, is piecewise linear. We have a set of $N$ disk-shaped robots initially centered at $\sta^1,\ldots,\sta^N$ with identical unit radii, and we denote $B(\bullet)$ as a unit circle centered at $\bullet$. The robots have distinct goal positions $\tar^{1,\cdots,N}$. Our method aims at finding a continuous path for each robot connecting $\sta^i$ and $\tar^i$, where the disk-shaped regions of any two robots do not overlap for any given time instance and no robot collides with obstacles.}

Having defined the problem, we provide an overview of our approach. The first step of our algorithm is pebble-graph embedding (\prettyref{sec:conversion}). The pebble-graph is embedded into $\WORK$ with the help of Blum's medial axis analysis, which can be performed through robust algorithms such as \cite{blum1978shape}. The medial axis analysis extracts two crucial pieces of information from a continuous workspace: skeleton lines and inscribed circles, as illustrated in \prettyref{fig:pipeline} (c). An inscribed circle, denoted as $\CIRCLE$, is defined as a circular subset of the workspace that touches the boundary of the workspace at least two points, and skeleton lines are derived by connecting centers of inscribed circles. These two pieces of information, skeleton lines and inscribed circles, are crucial because they allow the structure in $\WORK$ to move and accommodate robots: robots can reside in inscribed circles and move along sub-paths of skeleton lines. As illustrated in \prettyref{fig:pipeline} (d), we design an algorithm to select a subset of inscribed circles where robots can be arranged into connected loops such that they can move to nearby vacant positions or rotate along loops in a collision-free manner as illustrated in \prettyref{fig:pipeline} (e). By construction, our pebble-graph is topologically identified with the setting considered in \cite{yu2015pebble}. Our second step is motion planning (\prettyref{sec:discrete}), where we compute a series of motions to answer arbitrary MPP queries. In other words, robots can perform arbitrary position permutations restricted to the graph vertices. Finally, start and goal positions might not coincide with graph vertices, and we use unlabeled, local navigations to move robots between their start/goal positions and graph vertices (\prettyref{sec:navigation}).
\section{\label{sec:conversion}Pebble-Graph Embedding}
The first step of our method finds $M>N$ graph vertex positions using the greedy \prettyref{alg:greedy}. These vertices should be close to robots' start positions, so that robots can be moved to vertices with a high success rate via local navigation algorithms such as \cite{van2011reciprocal}. We further assume each robots' goal position set is close to its start position set, so goal positions can be ignore in the graph construct step. Our method maintains: 1) a set of considered robot start positions $\mathbb{X}$ initialized to be empty; 2) a set of remaining positions initialized to be $\{\sta^1,\cdots,\sta^N\}$; 3) a set of considered inscribed circles $\mathbb{C}$ initialized to be empty. During each iteration, an arbitrary position in the remaining set is picked, e.g. $\sta^i$. By the definition of inscribed circles, $B(\sta^i)$ must be contained in at least one inscribed circle, and we select the circle whose center is closest to $\sta^i$, denoted as $\CIRCLE(\sta^i)$. (To find $\CIRCLE(\sta^i)$, we sample the skeleton lines at regular intervals and extract inscribed circles centered at sample points, giving a discrete circle set $\bar{\mathbb{C}}$.) We then move $\sta^i$ from the remaining set to the considered set $\mathbb{X}$ and move $\CIRCLE(\sta^i)$ into the circle set $\mathbb{C}$. Further, if there is another $\sta^j$ in the remaining set such that $B(\sta^j)\subset\CIRCLE(\sta^i)$, we also move $\sta^j$ into the considered set. We terminate when the remaining set is empty. Note the success of our method depends on the order of choosing positions in the remaining set. \revised{To increase the success rate, we propose executing the algorithm multiple times using randomly shuffled start positions and return the first successful result. The resulting circle set would be used to construct our discrete loop graph using a BuildGraph function.}
\begin{rem}
The result of \prettyref{alg:greedy} is a graph $\SWAPGRAPH=\langle\VERTEX,\EDGE\rangle$ with $M$ vertices corresponding to $\VERTEX=\{\vertex^{1,\cdots,M}\}$ and edges $\EDGE$ constructed in the BuildGraph function. If $\SWAPGRAPH$ is not connected, has only one $loop$, or $M\leq N$, then we immediately return failure because our pebble motion planner (\prettyref{sec:discrete}) relies on these pre-conditions. Otherwise, we will use local navigation (\prettyref{sec:navigation}) to move robots from $\sta^{1,\cdots,N}$ to $N$ out of $M$ positions. If robots get stuck during local navigation, we also return failure.
\end{rem}
\begin{algorithm}[t]
\caption{\revised{Convert $\WORK$ into $\SWAPGRAPH$}}
\label{alg:greedy}
\begin{algorithmic}[1]
\State Perform Blum's medial axis analysis \cite{blum1978shape} for $\WORK$
\State A discrete candidate circle set $\bar{\mathbb{C}}$
\State Initialize considered set $\mathbb{X}\gets\emptyset$
\State Initialize selected circle set $\mathbb{C}\gets\emptyset$
\State Initialize $\SWAPGRAPH\gets<\emptyset,\emptyset>$
\label{ln:redo}
\For{$i=1,\cdots,N$}
\If{$\sta^i\notin\mathbb{X}$}
\State Find $\CIRCLE(\sta^i)\gets\argmin{B(\sta^i)\subseteq\CIRCLE\in\bar{\mathbb{C}}}\|\text{center}(\CIRCLE)-\sta^i\|$
\State $\mathbb{C}\gets\mathbb{C}\cup\{\CIRCLE(\sta^i)\}$
\State $\SWAPGRAPH\gets$BuildGraph($\mathbb{C}$)
\For{$j=1,\cdots,N$}
\If{$\sta^j\notin\mathbb{X}\land B(\sta^j)\subseteq\CIRCLE(\sta^i)$}
\State $\mathbb{X}\gets\mathbb{X}\cup\{\sta^j\}$
\EndIf
\EndFor
\EndIf
\EndFor
\If{Disconnected$(\SWAPGRAPH)\lor|\VERTEX|\leq N \lor \#\text{Loop}(\SWAPGRAPH)\leq 1$}
\If{\revised{Maximal trial number not reached}}
\label{ln:restartA}
\State \revised{Random shuffle $\E{s}^i$, $\mathbb{X}\gets\emptyset, \mathcal{G}\gets<\emptyset,\emptyset>$}
\State \revised{Goto} \prettyref{ln:redo}
\label{ln:restartB}
\Else
\State Return Failure
\EndIf
\Else
\State Return Success
\EndIf
\end{algorithmic}
\end{algorithm}

In the following, we analyze the relationship between loops and derive geometric conditions of a graph embedding, such that robot motions are collision-free. \revised{Since two loops can belong to different inscribed circles or different layers of the same inscribed circle, we use the subscript to index circles and superscript to index loops inside a circle. For example, $\mathcal{C}_a^i$ indicates the $i$th loop (starting from the innermost one) of the $a$th circle.}

\subsection{A Single Circle \label{sec:single_circle}}
We refine the details of the BuildGraph function, which arranges robots into loops within inscribed circles, ensuring that translational and rotational motions are collision-free. We denote the $i$th loop as $\CIRCLE^i\in\RR^2$, where the $0$th loop is the innermost loop. We use subscripts to distinguish different circles and superscripts to distinguish different loops. We have an upper bound on the number of loops $\CIRCLE$ can hold as $0\leq i\leq\floor*{(r(\CIRCLE)+1)/2}$, where $r(\CIRCLE)$ is the radius. In addition, we denote $\#(R)$ as the number of prescribed positions that can be put into some subset region $R\subseteq\RR^2$ without any collisions between positions or with $\partial\WORK$. 

We begin with the simplest case where there is only one $\CIRCLE$ in $\SWAPGRAPH$. Although a single robot can be put into $\CIRCLE^0$, we let $\#(\CIRCLE^0)=0$ because our pebble motion planner requires each loop to have at least 3 positions \revised{(see \prettyref{sec:discrete} for more details)}. For other loops, we have $\#(\CIRCLE^1)=6$ and for $i>1$ we have:

\footnotesize
\begin{align}
\label{eq:sizeLoop}
\#(\CIRCLE^i)=
\floor*{(2\pi-2\sin^{-1}\frac{1}{i})/
(2\sin^{-1}\frac{1}{2i})}+2.
\end{align}
\normalsize

\begin{wrapfigure}{r}{0.2\textwidth}
\vspace{-5px}
\includegraphics[width=0.2\textwidth]{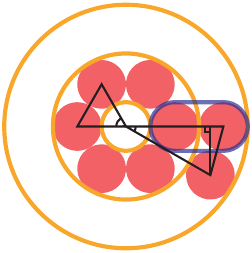}
%
\put(-65,53){\tiny{$60^\circ$}}
\put(-45,55){\tiny{$\sin^{-1}\frac{1}{i}$}}
\put(-40,39){\rotatebox{-30}{\tiny{$2i$}}}
\put(-22,45){\rotatebox{-90}{\tiny{$2$}}}
\vspace{-5px}
\caption{\small{\label{fig:sizeLoop} We illustrate the angles defining $\#(\CIRCLE^1)$ and $\#(\CIRCLE^i)$ for $i>1$. Our goal is to make sure the capsule-shaped region (blue) contains only the two robots.}}
\vspace{-10px}
\end{wrapfigure}
As illustrated in \prettyref{fig:sizeLoop}, \prettyref{eq:sizeLoop} allows a capsule-shaped region between $\CIRCLE^i$ and $\CIRCLE^{i-1}$ to contain only the two positions, allowing a pebble motion to be performed between the two loops. A pebble or rotational motion inside a single $\CIRCLE^i$ can be performed within $\CIRCLE^i$ without affecting any other loops by having all the positions trace out a circular arc along the centerline of $\CIRCLE^i$. In summary, the procedure to convert a single inscribed circle $\CIRCLE$ into a graph $\SWAPGRAPH$ is \prettyref{alg:single_circle}. Note that \prettyref{alg:single_circle} requires that $\floor*{(r(\CIRCLE)+1)/2}\geq2$, since our pebble motion planner requires more than one loop in $\SWAPGRAPH$. In addition, \prettyref{eq:sizeLoop} allows $\#(\CIRCLE^i)$ positions to be placed along the centerline of $\CIRCLE^i$ that are at least $2$ apart, positions of different layers are non-overlapping.

\begin{figure}[ht]
\centering
\includegraphics[width=0.95\linewidth]{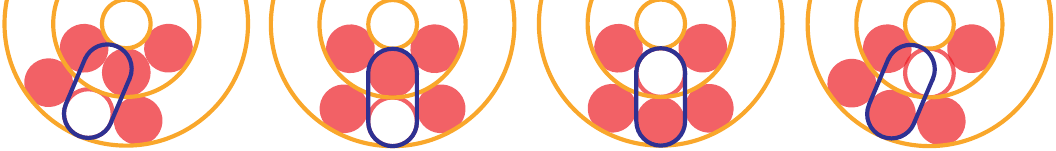}
\put(-125,5){\small{(a)}}
\put( -66,5){\small{(b)}}
\put(  -7,5){\small{(c)}}
\put( -207,15){\tiny{$\vertex $}}
\put( -217,6 ){\tiny{$\vertex'$}}
\put( -149,15){\tiny{$\vertex $}}
\put( -150,5 ){\tiny{$\vertex'$}}
\put( -90,15){\tiny{$\vertex'$}}
\put( -89,5 ){\tiny{$\vertex $}}
\put( -31,15){\tiny{$\vertex'$}}
\put( -38,6 ){\tiny{$\vertex $}}
\caption{\small{\label{fig:typeIISingle} To perform a pebble motion along $\vertex\leftrightarrow\vertex'$, we first move robots in $\CIRCLE^i$ continuously to: (a) align the capsule region (blue); (b) perform the swap; (c) move robots in $\CIRCLE^i$ backward.}}
\vspace{-5px}
\end{figure}
Each pebble or rotational motion in $\CIRCLE^i$ can be performed by moving robots along the centerline of $\CIRCLE^i$ without affecting other layers. To realize pebble motions between two robots $\vertex$ and $\vertex'$ of neighboring layers (\prettyref{ln:edgeI} of \prettyref{alg:single_circle}), we first rotate robots continuously in $\CIRCLE^i$ to make sure that the capsule-shaped region between $B(\vertex)$ and $B(\vertex')$ does not contain other robots, as illustrated in \prettyref{fig:typeIISingle}. After the pebble motion, we rotate robots in $\CIRCLE^i$ back to their original positions.
\begin{algorithm}[ht]
\caption{BuildGraph for a single $\CIRCLE$}
\label{alg:single_circle}
\begin{algorithmic}[1]
\If{$\floor*{\frac{r(\CIRCLE)+1}{2}}<2$}
\State Return $\SWAPGRAPH=<\emptyset,\emptyset>$
\EndIf
\For{$1\leq i\leq\floor*{\frac{r(\CIRCLE)+1}{2}}$}
\LineComment{Pick positions that are at least $2$ apart}
\State Pick $\#(\CIRCLE^i)$ positions along the centerline of $\CIRCLE^i$
\State Insert the $\#(\CIRCLE^i)$ positions into $\VERTEX$
\State Insert $\#(\CIRCLE^i)-1$ edges into $\EDGE$
\If{$i>1$}
\State Choose $\vertex,\vertex'$ belonging to $i,i-1$th loop\label{ln:edgeI}
\State Insert $\vertex\leftrightarrow\vertex'$ into $\EDGE$
\EndIf
\EndFor
\State Return $\SWAPGRAPH$
\end{algorithmic}
\end{algorithm}

\subsection{Two Overlapping Circles\label{sec:two_circle}}
\begin{figure}[ht]
\vspace{-5px}
\centering
\includegraphics[width=0.95\linewidth]{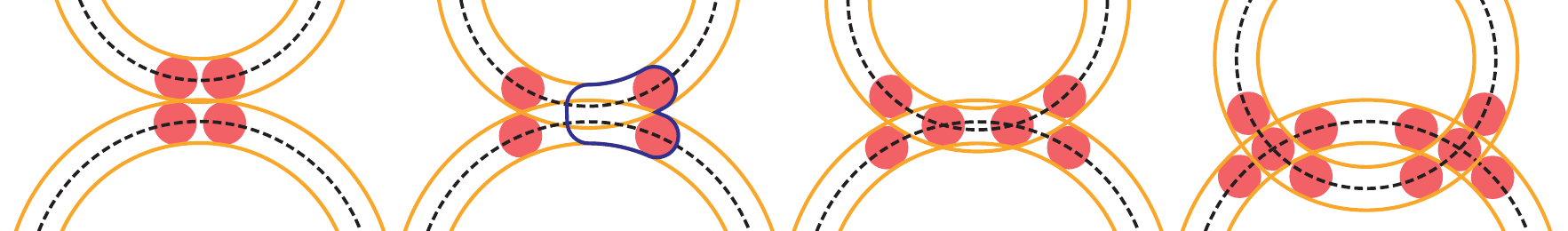}
\put(-208,30){\small{(a)}}
\put(-151,30){\small{(b)}}
\put(-93.5,30){\small{(c)}}
\put( -36,30){\small{(d)}}
\vspace{-5px}
\caption{\small{\label{fig:overlap} We illustrate 4 critical cases between two loops of overlapping inscribed circles.}}
\vspace{-10px}
\end{figure}

Next, we discuss the case where two circles $\CIRCLE_{a,b}$ are overlapping, which might result in loop sharing. Since the interiors of different layers in the same circle are disjointed, we always have the following decomposition of the domain:
\begin{align}
\label{eq:twoCircle}
\resizebox{.44\textwidth}{!}{
$\#(\CIRCLE_a\cup\CIRCLE_b)
\geq\sum_i\#(\CIRCLE_a^i-\CIRCLE_b)+
\sum_j\#(\CIRCLE_b^j-\CIRCLE_a)+
\sum_{i,j}\#(\CIRCLE_a^i\cap\CIRCLE_b^j)$,}
\end{align}
where the inequality can be strict since we exclude robots that can cross the boundary of sub-domains. We choose to construct our graph using the lower bound on the righthand side. As illustrated in \prettyref{fig:overlap}, the three terms on the righthand side can be derived analytically by considering 4 different cases. These four cases are distinguished by the distance between center vertices of $\CIRCLE_a$ and $\CIRCLE_b$, denoted as $D$. In addition, we require that the center of $\CIRCLE_a$ is outside $\CIRCLE_b$ and vice versa:
\begin{align}
\label{eq:circle_apart}
D\geq\max(r(\CIRCLE_a),r(\CIRCLE_b)).
\end{align}

\subsubsection{Case I}
The first case is illustrated in \prettyref{fig:overlap} (a), and it happens if the following condition holds:
\small
\begin{align}
\label{eq:cond1}
D\geq\sqrt{(2i+1)^2-1}+\sqrt{(2j+1)^2-1}
\end{align}
\normalsize
In this case, we cannot fit any circle in the overlapping area, i.e. $\#(\CIRCLE_a^i\cap\CIRCLE_b^j)=0$. The capacity of $\CIRCLE_a^i$ is reduced to:
\small
\begin{align}
\label{eq:capaCaseI}
\#(\CIRCLE_a^i-\CIRCLE_b^j)=
\floor*{\frac{2\pi-2\cos^{-1}\frac{D^2+(2i)^2-(2j+2)^2}{4iD}}
{2\sin^{-1}\frac{1}{2i}}}+1,
\end{align}
\normalsize
and a symmetric equation applies to $\CIRCLE_b^j$. Although the two loops overlap, they are not connected in $\SWAPGRAPH$ because $\CIRCLE_a^i\cap\CIRCLE_b^j$ is too narrow to perform pebble motions. It is trivial to show that rotational motions can be performed in either $\CIRCLE_a^i$ or $\CIRCLE_b^j$. We can convert this case by building two loops for $\CIRCLE_a^i$ and $\CIRCLE_b^j$ without adding edges to $\EDGE$.

\subsubsection{Case II}
The second case is illustrated in \prettyref{fig:overlap} (b), which happens if \prettyref{eq:cond1} does not hold but we have:
\begin{align}
\label{eq:cond2}
D\geq2i+2j.
\end{align}
In this case, we still have $\#(\CIRCLE_a^i\cap\CIRCLE_b^j)=0$ and \prettyref{eq:capaCaseI} holds. However, $\CIRCLE_a^i\cap\CIRCLE_b^j$ is now wide enough to allow an robot to travel in the blue region of \prettyref{fig:overlap} (b) to swap with a vacant position, which also utilizes the blue region. Therefore, we can insert an edge into $\EDGEI$ between two loops.

\subsubsection{Case III}
The third case is illustrated in \prettyref{fig:overlap} (c), which happens if \prettyref{eq:cond2} does not hold but we have:
\begin{align}
\label{eq:cond3}
D\geq2i+2j-2.
\end{align}
In this case, we have $\#(\CIRCLE_a^i\cap\CIRCLE_b^j)=2$ and we have:
\small
\begin{align*}
\#(\CIRCLE_a^i-\CIRCLE_b^j)=\text{(\prettyref{eq:capaCaseI})}-2.
\end{align*}
\normalsize
Moreover, the two robots in $\CIRCLE_a^i\cap\CIRCLE_b^j$ can be swapped if one of them is vacant and rotational motions can be performed in either $\CIRCLE_a^i$ or $\CIRCLE_b^j$. Therefore, we can construct two loops for $\CIRCLE_a^i$ and $\CIRCLE_b^j$, let them share two robots in $\CIRCLE_a^i\cap\CIRCLE_b^j$, and the edge between them.

\subsubsection{Case IV}
The last case is illustrated in \prettyref{fig:overlap} (d), and it happens if we have:
\begin{align}
\label{eq:cond4}
\max(r(\CIRCLE_a),r(\CIRCLE_b))\leq D<2i+2j-2.
\end{align}
In this case, we have $\#(\CIRCLE_a^i\cap\CIRCLE_b^j)=2$, but $\#(\CIRCLE_a^i-\CIRCLE_b)$ has a new expression:
\begin{align}
\label{eq:capaCaseIV}
\resizebox{.42\textwidth}{!}{
$\#(\CIRCLE_a^i-\CIRCLE_b^j)=
\floor*{\frac{2\cos^{-1}\frac{D^2+(2i)^2-(2j-2)^2}{4iD}}
{2\sin^{-1}\frac{1}{2i}}}+1+
\text{\prettyref{eq:capaCaseI}}$.}
\end{align}
Similar to case III, we can construct two loops for $\CIRCLE_a^i$ and $\CIRCLE_b^j$, let them share two robots in $\CIRCLE_a^i\cap\CIRCLE_b^j$, but the two loops do not share any edges.

\revised{Note that we have computed $\#(\CIRCLE_a^i-\CIRCLE_b^j)$ in the four cases but we need $\#(\CIRCLE_a^i-\CIRCLE_b)$ in \prettyref{eq:twoCircle}, which can be computed by excluding each layer $\CIRCLE_b^j$ from $\CIRCLE_a^i$ (the details are similar to the four cases and omitted for brevity). We summarize our method to convert $\WORK$ to $\SWAPGRAPH$ for two overlapping circles in \prettyref{alg:two_circle}. This algorithm inserts vertices corresponding to each term of the righthand side of \prettyref{eq:twoCircle} in \prettyref{ln:IntersectArea} and \prettyref{ln:CircleArea}, respectively, and then inserts edges to connect loops. Note that we only need to insert inter-loop edges in Case II, as is done in \prettyref{ln:InterLoopEdge}.}

\begin{algorithm}[ht]
\caption{BuildGraph for two circles $\CIRCLE_{a,b}$}
\label{alg:two_circle}
\begin{algorithmic}[1]
\For{$1\leq i\leq\floor*{\frac{r(\CIRCLE_a)+1}{2}}$ and 
$1\leq j\leq\floor*{\frac{r(\CIRCLE_b)+1}{2}}$}
\State Insert $\#(\CIRCLE_a^i\cap\CIRCLE_b^j)$ vertices into $\VERTEX$
\label{ln:IntersectArea}
\EndFor
\For{pass=$1,2$}
\For{$1\leq i\leq\floor*{\frac{r(\CIRCLE_a)+1}{2}}$}
\State Insert $\#(\CIRCLE_a^i-\CIRCLE_b)$ vertices into $\VERTEX$\label{ln:CircleArea}
\State Add $\#(\CIRCLE_a^i-\CIRCLE_b)+\sum_j\#(\CIRCLE_a^i\cap\CIRCLE_b^j)$ edges to $\EDGE$
\If{$i>1$}
\State Choose $\vertex,\vertex'$ belonging to $i,i-1$th loop
\State Insert $\vertex\leftrightarrow\vertex'$ into $\EDGE$
\EndIf
\EndFor
\State Swap $a,b$
\EndFor
\For{$1\leq i\leq\floor*{\frac{r(\CIRCLE_a)+1}{2}}$ and 
$1\leq j\leq\floor*{\frac{r(\CIRCLE_b)+1)}{2}}$}
\If{Case II holds}
\State Choose $\vertex,\vertex'$ from $i,j$th loop of $\CIRCLE_{a,b}$, respectively
\State Insert $\vertex\leftrightarrow\vertex'$ into $\EDGE$\label{ln:InterLoopEdge}
\EndIf
\EndFor
\State Return $\SWAPGRAPH$
\end{algorithmic}
\end{algorithm}
Informally, we justify the correctness of \prettyref{alg:two_circle}. First, since each summand in \prettyref{eq:twoCircle} corresponds to pairwise disjointed regions, the embedding is collision-free. Second, it is trivial to show that pebble or rotational motions within a single circle can be performed in the same way as in \prettyref{sec:single_circle}. In addition, pebble motions between two overlapping loops are only required in Case II, which can also be safely performed, as shown in \prettyref{fig:overlap} (b). Note that we might not get a valid pebble graph in two scenarios: 1) when some loop might not have 3 vertices; 2) when two circles are not connected, leading to disconnected $\SWAPGRAPH$.

\subsection{More Than Two Circles}
We can combine the two previous cases to derive the final BuildGraph procedure. We assume that $K$ inscribed circles $\CIRCLE_{1,\cdots,K}$ are used and our algorithm is based on the assumption that, for any three (pairwise distinct) circles $\CIRCLE_a,\CIRCLE_b,\CIRCLE_c$, we have:
\begin{align}
\label{eq:assumption}
\CIRCLE_a\cap\CIRCLE_b\cap\CIRCLE_c=\emptyset.
\end{align}
As a result, we have the following inequality:
\begin{align}
\label{eq:multiCircle}
\resizebox{.43\textwidth}{!}{
$\#(\bigcup_{a=1}^K\CIRCLE_a)
\geq\sum_{a,i}\#(\CIRCLE_a^i-\bigcup_{b\neq a}\CIRCLE_b)+
\sum_{b<a}\sum_{i,j}\#(\CIRCLE_a^i\cap\CIRCLE_b^j)$,}
\end{align}
\begin{wrapfigure}{r}{0.2\textwidth}
\vspace{-5px}
\includegraphics[width=0.9\linewidth]{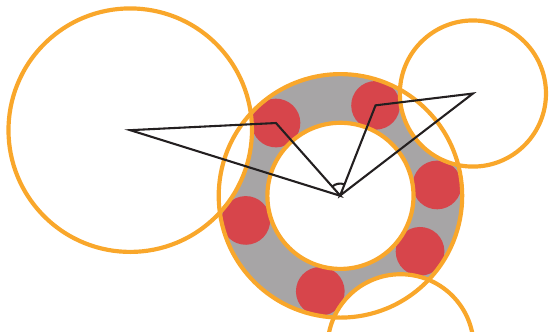}
\put(-38,27){\small{$\theta$}}
\caption{\small{\label{fig:residual_volume} We illustrate the procedure to compute $\#(\CIRCLE_a^i-\bigcup_{b\neq a}\CIRCLE_b)$.}}
\vspace{-5px}
\end{wrapfigure}
which is valid only when \prettyref{eq:assumption} holds. We can compute each of the terms in \prettyref{eq:multiCircle} analytically. For a term of type $\#(\CIRCLE_a^i\cap\CIRCLE_b^j)$, we can compute it using the four cases in \prettyref{sec:two_circle}. For a term of type $\#(\CIRCLE_a^i-\bigcup_{b\neq a}\CIRCLE_b)$, we use a procedure illustrated in \prettyref{fig:residual_volume}, where we first identify all the tangent cases (red circle) using triangular relationships (black), then find the angles between tangent cases ($\theta$), and finally compute the number of spheres that can be put into the interval between neighboring tangent cases as $\floor*{\theta/(2\sin^{-1}1/(2i))}+1$.
\begin{algorithm}[ht]
\caption{BuildGraph for $\CIRCLE_{1,\cdots,K}$}
\label{alg:multi_circle}
\begin{algorithmic}[1]
\For{$1\leq a<b\leq K$}
\For{$1\leq i\leq\floor*{\frac{r(\CIRCLE_a)+}{2}}$ and 
$1\leq j\leq\floor*{\frac{r(\CIRCLE_b)+1}{2}}$}
\State Insert $\#(\CIRCLE_a^i\cap\CIRCLE_b^j)$ positions into $\VERTEX$\label{ln:IntersectAreaMore}
\EndFor
\EndFor
\For{$1\leq a\leq K$}
\For{$1\leq i\leq\floor*{\frac{r(\CIRCLE_a)+1}{2}}$}
\State Insert $\#(\CIRCLE_a^i-\bigcup_{b\neq a}\CIRCLE_b)$ positions into $\VERTEX$\label{ln:CircleAreaMore}
\State Insert 
$\#(\CIRCLE_a^i-\bigcup_{b\neq a}\CIRCLE_b)+\sum_{b\neq a}\sum_j\#(\CIRCLE_a^i\cap\CIRCLE_b^j)$
\State edges into $\EDGE$\label{ln:EdgeTypeI}
\If{$i>1$}
\State Choose $\vertex,\vertex'$ belonging to $i,i-1$th loop
\State Insert $\vertex\leftrightarrow\vertex'$ into $\EDGE$\label{ln:EdgeTypeII}
\EndIf
\EndFor
\EndFor
\For{$1\leq a<b\leq K$}
\For{$1\leq i\leq\floor*{\frac{r(\CIRCLE_a)+1}{2}}$ and 
$1\leq j\leq\floor*{\frac{r(\CIRCLE_b)+1}{2}}$}
\If{Case II holds}
\State Choose $\vertex,\vertex'$ from $i,j$th loop of $\CIRCLE_{a,b}$
\State Insert $\vertex\leftrightarrow\vertex'$ into $\EDGE$\label{ln:EdgeTypeIII}
\EndIf
\EndFor
\If{There is $\tau$ satisfying \prettyref{eq:path}}
\State Set $i=\floor*{\frac{r(\CIRCLE_a)+1}{2}}$ and $j=\floor*{\frac{r(\CIRCLE_b)+1}{2}}$
\State Choose $\vertex,\vertex'$ from $i,j$th loop of $\CIRCLE_{a,b}$
\State Insert $\vertex\leftrightarrow\vertex'$ into $\EDGE$\label{ln:EdgeTypeIV}
\EndIf
\EndFor
\State Return $\SWAPGRAPH$
\end{algorithmic}
\end{algorithm}

\begin{figure}[ht]
\vspace{-5px}
\centering
\includegraphics[width=0.95\linewidth]{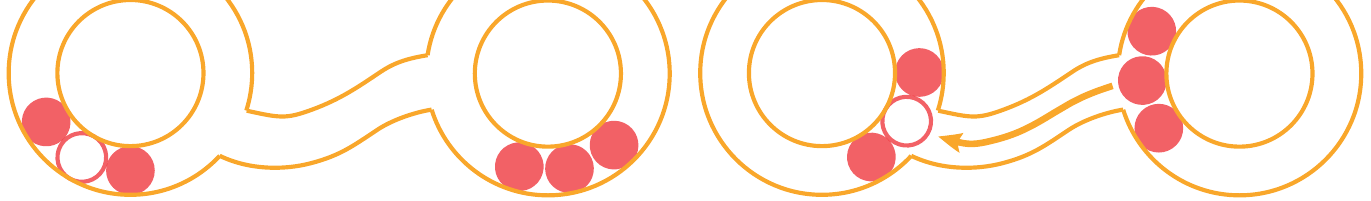}
\put(-216,20){\small{(a)}}
\put(-99,20){\small{(b)}}
\vspace{-5px}
\caption{\small{\label{fig:pathAxis} A tunnel along the medial axis is added between two distant loops (a), so that an agent can be moved to a nearby vacant position by first rotating in the two loops and then moving the agent along the tunnel (b).}}
\vspace{-10px}
\end{figure}
However, as shown in \prettyref{sec:two_circle}, two scenarios might lead to invalid pebble-graphs that also apply for multiple circles. First, there may be invalid loops with less than 3 positions. In \prettyref{sec:two_circle}, we eliminate this case by having two circles' centers outside each other, but this cannot be done for multiple circles. Second, two circles might be too far apart, leading to disconnected $\SWAPGRAPH$. This later scenario can be mitigated with the help of a medial axis. For two circles $\CIRCLE_{s,t}$, their centers are on the medial axis. If we can find a sub-path $\tau: [0,1]\rightarrow\RR^2$ along the medial axis such that:
\small
\begin{equation}
\begin{aligned}
\label{eq:path}
&\tau(0)\in\CIRCLE_a\land\tau(1)\in\CIRCLE_b\land\\
&(\tau\oplus B(0))-(\CIRCLE_s\cup\CIRCLE_t)\subseteq\WORK-\bigcup_{a=1}^K\CIRCLE_a,
\end{aligned}
\end{equation}
\normalsize
then we can insert an edge into $\EDGEI$ between the outermost loops of $\CIRCLE_s$ and $\CIRCLE_t$. Here $\oplus$ is the Minkowski Sum, i.e. we require the path to have no interference with any obstacle or other circle. When performing pebble motions along $\tau$, we follow the procedure illustrated in \prettyref{fig:pathAxis}. The motions can be performed in a collision-free manner when \prettyref{eq:path} holds. \revised{ The procedure to convert the $K$ circles into $\SWAPGRAPH$ is summarized in \prettyref{alg:multi_circle}. This algorithm adds vertices corresponding to each term of the righthand side of \prettyref{eq:assumption} in \prettyref{ln:IntersectAreaMore} and \prettyref{ln:CircleAreaMore}, respectively. It then adds three kinds of edges: 1) loop edges (\prettyref{ln:EdgeTypeI}); 2) inter loop edges (\prettyref{ln:EdgeTypeII} and \prettyref{ln:EdgeTypeIII}); and 3) medial axis edges (\prettyref{ln:EdgeTypeIV}).}
\section{\label{sec:discrete}Pebble Motion Planning}
We allow robots to perform two types of movements on the pebble graph: 1) cyclic permutation of robots in a single loop; 2) movement of robot to an adjacent vacant position (either in the same loop or in an adjacent loop). This setting is similar to prior work in \cite{Yu2015OptimalMP}, but that method focuses on checking the feasibility, while we ensure feasibility of arbitrary robot configuration change on the graph under the assumption that the graph vertices are not fully occupied.  \revised{Our pebbling motion planner does not differentiate between loops of the same circle or loops of different circles (because such differences have been taken care of in \prettyref{sec:conversion}). Therefore, we only use a global superscript to index loops. For the $i$th loop, the set of vertices is denoted as $\VERTEXL^i$. These vertices are connected by a set of edges denoted as $\EDGEL^i$.}

Formally, the topology of a pebble-graph $\SWAPGRAPH=<\VERTEX,\EDGE>$ successfully constructed from \prettyref{alg:greedy} is simple, undirected, and connected. Different loops can be interpreted as a partition of $\VERTEX$ as follows:
\begin{align}
\label{eq:vpart}
\VERTEX=\bigcup_{i=1}^K\VERTEXL^i,\quad
|\VERTEXL^i\cap\VERTEXL^j|\in\{0,2\}.
\end{align}
The partition of vertices induces a partition of edge set $\EDGE$ as follows:
\begin{small}
\begin{align}
\label{eq:epart}
\EDGE=\EDGEI \cup \bigcup_{i=1}^K\EDGEL^i,\quad
\EDGEL^i\cap\EDGEI=\emptyset,\quad
|\EDGEL^i\cap\EDGEL^j|\in\{0,1\}.
\end{align}
\end{small}
Specifically, we assume that the vertices can be partitioned into $K>1$ loops, where $K$ is the number of loops in the graph. We also require that $\EDGEL^i$ is the unique loop connecting $\VERTEXL^i$ and that $\EDGEL^i$ is a simple loop (with no repeated vertices). Note that, under these definitions, each loop must have at least $3$ vertices, i.e. $|\VERTEXL^i|\geq3$. This is necessary because our motion planner moves robots by swapping the positions of two neighbors connected by a graph edge, where we need a third position as a buffer to accommodate temporary robots (see \ifsupp\prettyref{sec:proof} \else our extended version \cite{2002.11892} \fi for more details). Finally, we allow two loops to overlap; however, if the $i$th loop and the $j$th loop overlap, we require that they share exactly $2$ vertices. As a result, two loops can also share a common edge (this is why we have $|\VERTEXL^i\cap\VERTEXL^j|\in\{0,2\}$ and $|\EDGEL^i\cap\EDGEL^j|\in\{0,1\}$). This corresponds to Case III of \prettyref{sec:two_circle}. In summary, a successful graph returned by \prettyref{alg:greedy} satisfies the following conditions:
\begin{defi}[Pebble Graph]
\label{def:swap}
A pebble graph $\SWAPGRAPH$ is a simple, undirected, and connected graph satisfying \prettyref{eq:vpart} and \prettyref{eq:epart} with $K>1$ such that, through each group of vertices in $\VERTEXL^i$ ($i=1,\cdots,K$), there is a simple, closed path formed by edges in $\EDGEL^i$.
\end{defi}
On such a graph, we could follow similar reasoning as \cite{auletta1999linear,yu2015pebble} to verify feasibility for all the MPP instances and construct the corresponding motion plans, as summarized in the following result:
\begin{thm}[Pebble-Graph Feasibility]
\label{thm:complete}
On a pebble-graph $\SWAPGRAPH$ with $|\VERTEX|=M>N$, we assume $\sta^i=\vertex^i$ and $\tar^i=\vertex^\sigma(i)$, where $\sigma(\bullet)$ is an arbitrary permutation. there exists a finite sequence of pebble or rotational motions to move robots from $\sta^i$ to $\tar^i$ for all $i=1,\cdots,N$, and the length of the motion sequence is $\mathcal{O}(|\VERTEX|^2)$.
\end{thm}
We prove \prettyref{thm:complete} constructively in \ifsupp\prettyref{sec:proof} \else our extended version \cite{2002.11892} \fi by converting the permutation into a sequence of pairwise position swaps between two neighboring vertices, and we use the proof to construct a planning algorithm to solve MPP instances in \ifsupp\prettyref{sec:planning}. \else our extended version \cite{2002.11892}.\fi We then show that each swap can be accomplished by a $\mathcal{O}(|\VERTEX|)$-sequence of motions. We further show that the amortized length of motion sequence for permuting the position of two arbitrarily distant vertices is also $\mathcal{O}(|\VERTEX|)$. Therefore, the total length of motion sequence to solve the pebbling problem is $\mathcal{O}(|\VERTEX|^2)$. To accomplish each position swap, we need to move the vacant vertex, which is not occupied by any robot, near the to-be-swapped vertices. We could optimize the motion plan by moving the vacant vertex along the shortest path. It is convenient to pre-compute the all-pair shortest paths, which costs $\mathcal{O}(|\VERTEX|^3)$. This step dominates the complexity of computing the motion sequence.
\begin{figure*}[ht]
\vspace{-5px}
\centering
\includegraphics[width=0.22\linewidth]{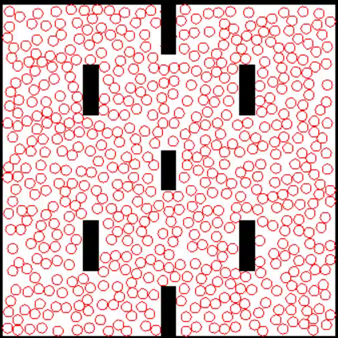}
\includegraphics[width=0.22\linewidth]{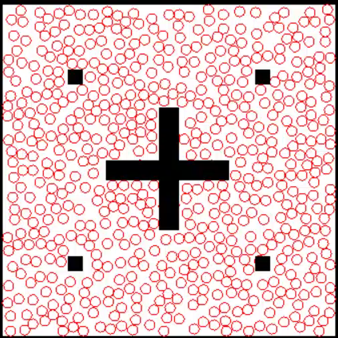}
\includegraphics[width=0.22\linewidth]{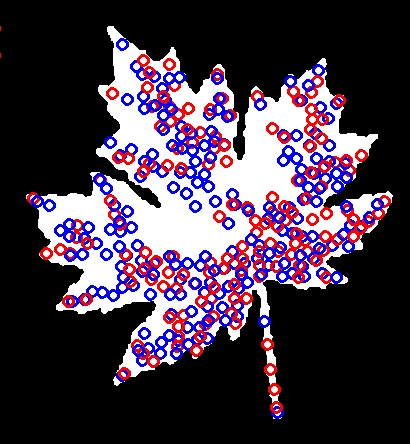}
\includegraphics[width=0.22\linewidth]{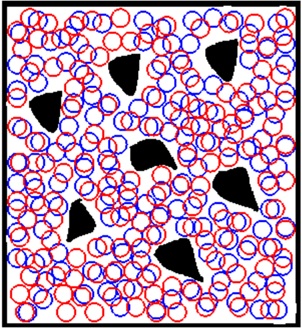}
\put(-450,5){(a)}
\put(-325,5){(b)}
\put(-200,5){\textcolor{white}{(c)}}
\put( 5  ,5){(d)}
\caption{\label{fig:bench}  \revised{We highlight the $4$ most challenging benchmarks. (a): A rectangular workspace with $600$ robots; (b): Another obstacle setup with $710$ robots. The two benchmarks (ab) are also used in \cite{yu2018effective}, where robots' goal positions are derived by permuting their start positions. (c): A maple-shaped workspace with highly irregular boundaries and $100$ robots; (d): A rectangular workspace with irregular obstacles and $130$ large robots. For (c) and (d), robots' start positions are in red and goal positions are in blue.}}
\vspace{-15px}
\end{figure*}
\begin{figure}[ht]
\centering
\includegraphics[width=0.80\linewidth]{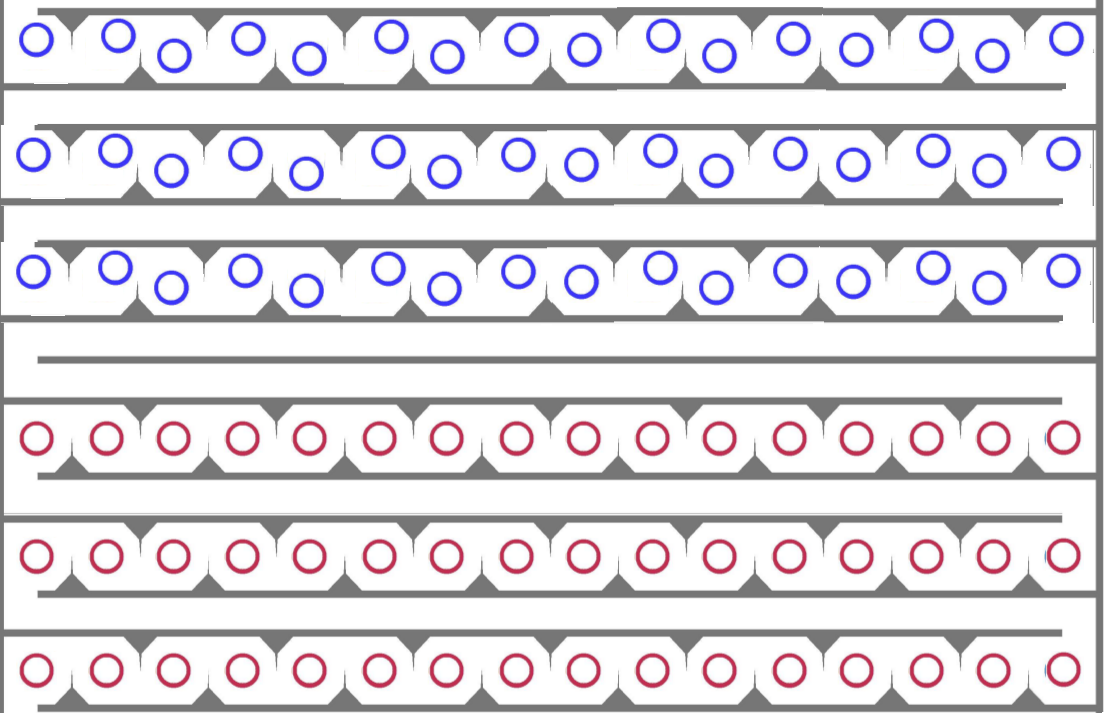}
\caption{A similar scenario to one used in~\cite{solomon2018motion}, where robots need to give way to each other by revolving. The robots' start and end positions are in red and blue, respectively.}
\vspace{-5px}
\end{figure}
\begin{figure}[ht]
\centering
\vspace{-10px}
\includegraphics[width=0.85\linewidth]{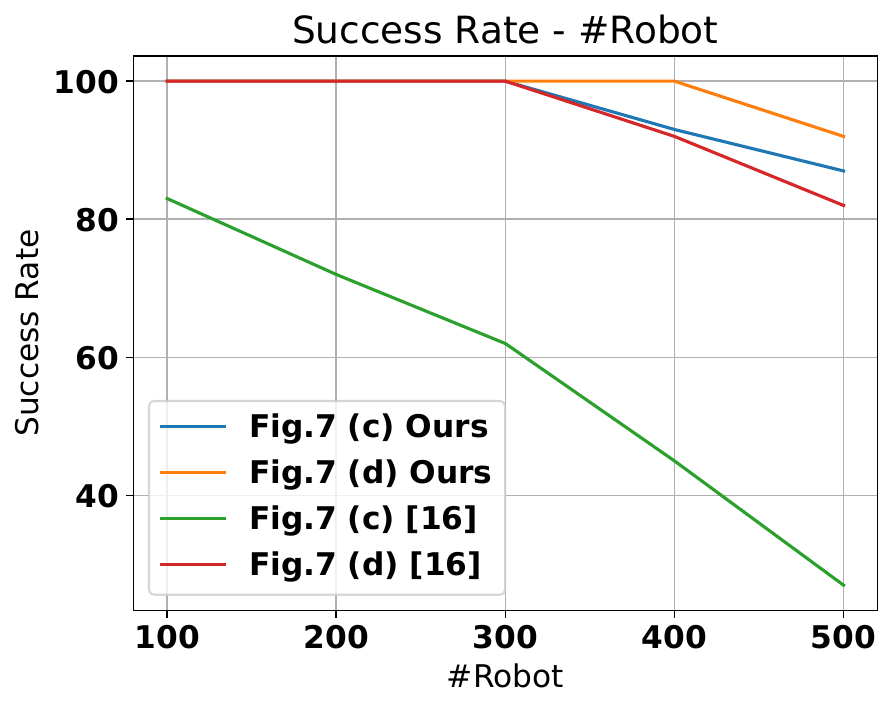}
\caption{\label{fig:trial}  \revised{We compare the changes in success rate over $50$ random runs of our method and \cite{yu2018effective}, under different robot densities. In cases of high density, our method exhibits a much higher success rate, especially with a narrow passage, as seen in Figure 7 (c).}}
\vspace{-10px}
\end{figure}
\begin{table}[H]
\resizebox{\linewidth}{!}{%
\begin{tabular}{c|c|cccccc}
\rowcolor{Gray}
\toprule
Bench. & \diagbox{Method}{Metric} & Precomp. & Planning & Total & Succ. & Makespan & Traj. Length     \\ 
\midrule
\multirow{3}{*}{\prettyref{fig:bench} (a)/$600$} & Ours & 
2 & 313 & 316 & 100\% & 73$\times$600 & 178$\times$600    \\
 & \cite{yu2018effective} & 
- & 333 & 333 & 100\% & 81$\times$600 & 171$\times$600    \\
 & \cite{solomon2018motion} & 
- & 258 & 258 & - & - & -   \\
\midrule
\multirow{3}{*}{\prettyref{fig:bench} (b)/$710$} & Ours & 
2 & 451 & 453 & 100\% & 82$\times$710 & 163$\times$710    \\
 & \cite{yu2018effective} & 
- & 432 & 432 & 100\% & 76$\times$710 & 166$\times$710    \\
 & \cite{solomon2018motion} & 
- & 357 & 357 & - & - &-    \\
\midrule
\multirow{3}{*}{\prettyref{fig:bench} (c)/$170$} & Ours & 
4 & 212 & 216 & 100\% & 33$\times$170 & 122$\times$170    \\
 & \cite{yu2018effective} & 
- & 208 & 208 & 82\% & 28$\times$170 & 118$\times$170    \\
 & \cite{solomon2018motion} & 
- & 205 & 205 & - & - & -    \\
\midrule
\multirow{3}{*}{\prettyref{fig:bench} (d)/$130$} & Ours & 
3 & 128 & 131 & 100\% & 26$\times$130 & 132$\times$130    \\
 & \cite{yu2018effective} & 
- & 122 & 122 & 100\% & 23$\times$130 & 136$\times$130    \\
 & \cite{solomon2018motion} & 
- & 107 & 107 & - & - & -   \\
\bottomrule
\end{tabular}}
\vspace{5px}
\caption{\label{table:profiling}\revised{ We compare the performance of different techniques. From left to right: benchmark/robots number, algorithm name, time to construct the pebble graph (in seconds); time to compute the MPP motion plan (in seconds); total computation time (in seconds); rate of success; makespan (average makespan of each robot $\times$ number of robots); trajectory length (average trajectory length of each robot $\times$ number of robots). All numbers are averaged over 50 random trials. Note the makespan (measured in the number of pebble steps) is incomparable with the trajectory length (measured in the unit length traveled by each robot).}}
\vspace{-5px}
\end{table}
\section{\label{sec:navigation}Local Navigation}
In general, robot start/goal positions do not coincide with the graph vertices and we use local navigations to move robots to/from graph vertices. To this end, we propose a heuristic method based on RVO \cite{van2011reciprocal} and CAPT \cite{6630671}. The RVO algorithm resolves local collisions between robots, and CAPT further uses the Hungarian algorithm to solve unlabeled navigation problems by assigning robots to goal positions. Since our graph vertices do not coincide with robot start/goal positions, we use a similar technique to assign each $\sta^i/\tar^i$ to some graph vertex. This assignment can be arbitrary in our method because robots can be moved to any graph vertices and permuted later, while we propose computing an as-close-as-possible assignment via optimal transport by solving the following mixed integer linear programming:
\small
\begin{align*}
\argmin{z_\sta^{ij}\in\{0,1\}}\;&\sum_{i=1}^N\sum_{j=1}^Mz_\sta^{ij}\|\sta^i-\vertex^j\| \\
\text{s.t.}\;&\sum_{j=1}^Nz_\sta^{ij}=1\wedge\sum_{i=1}^Nz_\sta^{ij}=1,
\end{align*}
\normalsize
where $z_\sta^{ij}=1$ implies assigning $\sta^i$ to $\vertex^j$. After the assignment is computed, we can move each $\sta^i$ to $\vertex^j$ using RVO. An identical procedure is used to assign $\tar^i$ to $\vertex^j$ with decision variables denoted as $z_\tar^{ij}$. Finally, the graph vertex permutation can be determined from $z_\sta^{ij}$ and $z_\tar^{ij}$. We set $\sigma(j)=j'$ if $z_\sta^{ij}=1$ and $z_\tar^{ij'}=1$ for some $i$.
\section{\label{sec:experiments}Experiments}
\revised{We evaluate the performance of our method on a set of $5$ benchmarks. The algorithm is implemented in C++ and tested on a desktop machine with an Intel Core i7 CPU running at 3.30GHz with 16GB of RAM. We have also compared our algorithm with~\cite{yu2018effective,solomon2018motion}, which are recently proposed methods for centralized motion planning in continuous workspaces. Our method can compute motion plans for up to $200$ robots in less than $10$ seconds shown in \prettyref{fig:bench} (a) and (b), where the cost of pebble-graph embedding (\prettyref{alg:greedy}) is marginal and the majority of computation is spent on scheduling pebble motions on the graph. 
The detailed timing and trajectory qualities of the three methods are summarized in \prettyref{table:profiling}. These results are derived by performing 50 random trials and taking the average. For each random trial, the robots' start positions are randomly sampled.}

\revised{Our first benchmark is illustrated in \prettyref{fig:bench} (a), where we compare our method and prior work \cite{yu2018effective} under different robot densities. The robots' start positions are shown in red and their goal positions are derived by randomly permuting the starts. For $100$ robots, our method takes $8$ seconds to compute the motion plan (including pebble-graph construction and pebble motion planning, but not local navigation), while it takes $10$ seconds to compute the motion plan using \cite{yu2018effective}. We then increase the number of robots to $200$ in \prettyref{fig:bench} (a), where our algorithm still takes $8$ seconds to find a motion plan, while the computational cost of \cite{yu2018effective} is $18$ seconds using a rectangular grid as the graph. We can further increase the number of robots up to $600$, resulting in robots occupying $47\%$ of $|\WORK|$, and the motion plan can be computed within $313$ seconds. We have tried a similar benchmark (\prettyref{fig:bench} (b)) with a different obstacle setup, where the overall computation time of our method is $2$ seconds for $70$ robots. We can increase the number of robots up to $710$, occupying $55\%$ of $|\WORK|$, for which our method computes a motion plan within $453$ seconds. Our third benchmark in \prettyref{fig:bench} (c) involves a maple-shaped boundary with a narrow passage and our forth benchmark contains irregular obstacles. Both~\cite{yu2018effective,solomon2018motion} fail for these irregular cases. This is because the regular grid used by \cite{yu2018effective} does not fit into the narrow space and the well-separated assumption of \cite{solomon2018motion} does not hold,  In contrast, our method succeeds in computing a motion plan within $220$ seconds for $170$ robots in \prettyref{fig:bench} (c) and $130$ robots in \prettyref{fig:bench} (d), with a success rate of $100\%$ over all $20$ executions.}

Furthermore, we implement a similar scenario to \cite{solomon2018motion}, but unlike their experiments, the robots in our experiments are randomly placed. As a results, some robots do not have sufficient free space to accomplish a revolving motion, which is needed in \cite{solomon2018motion} to find feasible motion plans. However, our method successfully moves robots into inscribed circles via local navigation, and revolving motions can still be performed. These behaviors are illustrated in the video and the overall computation time is less than $3$ seconds.

In \prettyref{fig:trial}, we have profiled the success rate of our method and \cite{yu2018effective} under different scenarios and robot densities. Note that our method relies on a randomized, greedy \prettyref{alg:greedy} to construct the pebble graph, so it is possible to improve our success rate by running \prettyref{alg:greedy} multiple times using different random seeds until a solution is found, as in \prettyref{ln:restartA} to \prettyref{ln:restartB} of \prettyref{alg:greedy}. 
\ifsupp
\begin{figure*}[ht]
\centering
\includegraphics[width=0.8\linewidth]{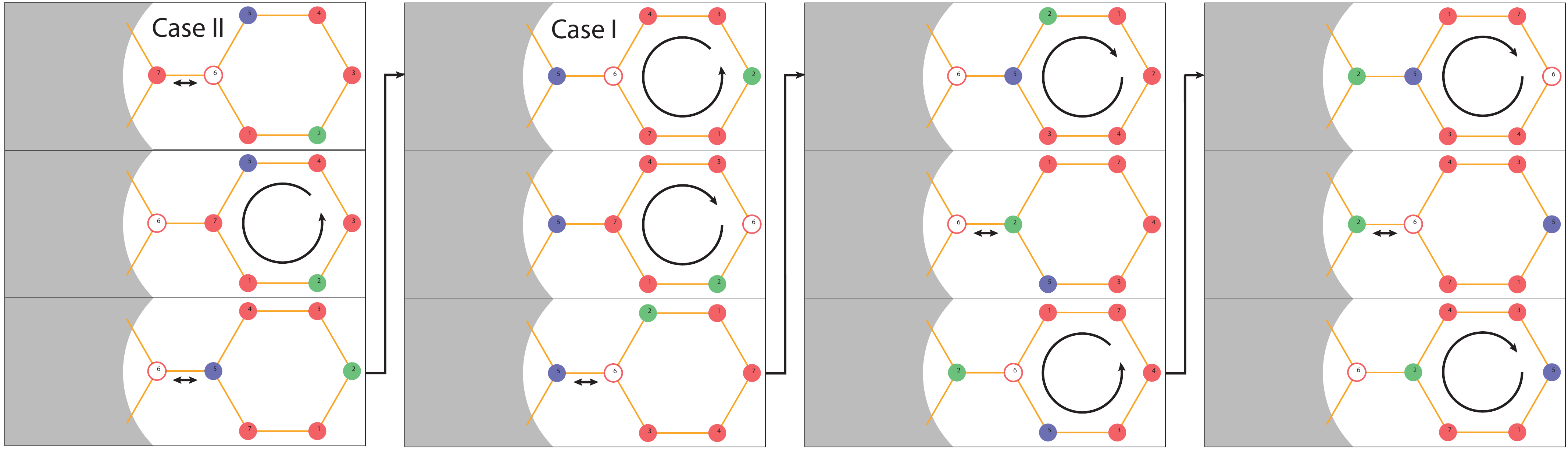}
\caption{\small{\label{fig:CaseI_II} We illustrate Case I and Case II of a position swap. Case II can be reduced to Case I via three additional moves. These moves only affect the vertices within one loop and a connected vertex in another loop, which can always be found because we assume $K>1$. Note that the motion sequence of Case I does not affect other vertices than the to-be-swapped ones. (All the vertices are labeled, zoom in to view.)}}
\vspace{-20px}
\end{figure*}
\fi
\section{\label{sec:conclusion}Conclusion and Limitations}
We presented a new method to bridge the gap between continuous MPP planning and discrete pebble-graph motion. We first use medial axis analysis to extract critical information from the workspace, i.e. skeleton lines and inscribed circles. Using this information, we convert the free space into a pebble-graph via embedding. 
\begin{wrapfigure}{r}{0.2\textwidth}
\includegraphics[width=0.98\linewidth]{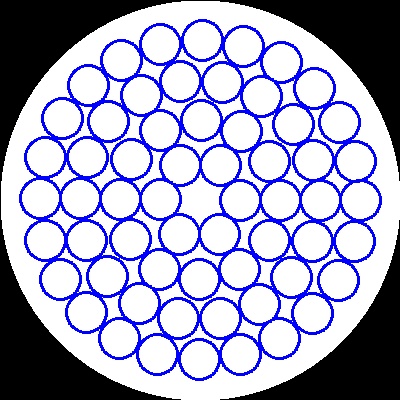}
\caption{\small{\label{fig:extreme} In the extreme case, densely packed robots take up $68\%$ of $\WORK$.}}
\vspace{-10px}
\end{wrapfigure}
We show that motion planning on the pebble-graph is always feasible under mild assumptions and general MPP instances can be reduced to graph pebbling problems via local navigation. We conduct experiments using a set of $5$ challenging benchmarks and achieve a $100\%$ success rate under robot densities less than $37\%$. In \prettyref{fig:extreme}, the robots take up $68\%$ of the free space, which implies that our method can work under extreme robot densities. The major limitation of our method lies in the overly conservative conditions derived in \prettyref{sec:conversion} which can leave some gap regions between robots. In large open areas, regular grids can have better space coverage \cite{yu2018effective}. Our future work would consider hybrid graph embedding techniques that combine multiple space tiling patterns, and we plan to derive improved boundary conditions for neighboring inscribed circles to accommodate more robots. Another limitation is that, our \prettyref{alg:greedy} only considers robots' start positions, and ignores their goals. This works well if the goal set is derived by permuting the start positions (as in \prettyref{fig:bench} (ab)), but the local navigation can fail if goal sets are far from the start positions (as in \prettyref{fig:bench} (cd)).
\ifsupp
\section{\label{sec:proof}Appendix: Pebble-Graph Feasibility}
We denote by $\vacant$ as the graph vertex that is not occupied by any robot. The following result is obvious.
\begin{lem}[Move of $\vacant$]
\label{lem:vacantMove}
On $\SWAPGRAPH$ with $|\VERTEX|=N+1$, the position of $\vacant$ can be moved to any vertex.
\end{lem}
\begin{proof}
Since $\SWAPGRAPH$ is connected, there must be a path connecting $\vacant$ and $\vertex$. We can then move $\vacant$ to $\vertex$ by pebble motions along the path.
\end{proof}
\begin{wrapfigure}{r}{0.2\textwidth}
\vspace{-15px}
\includegraphics[width=0.98\linewidth]{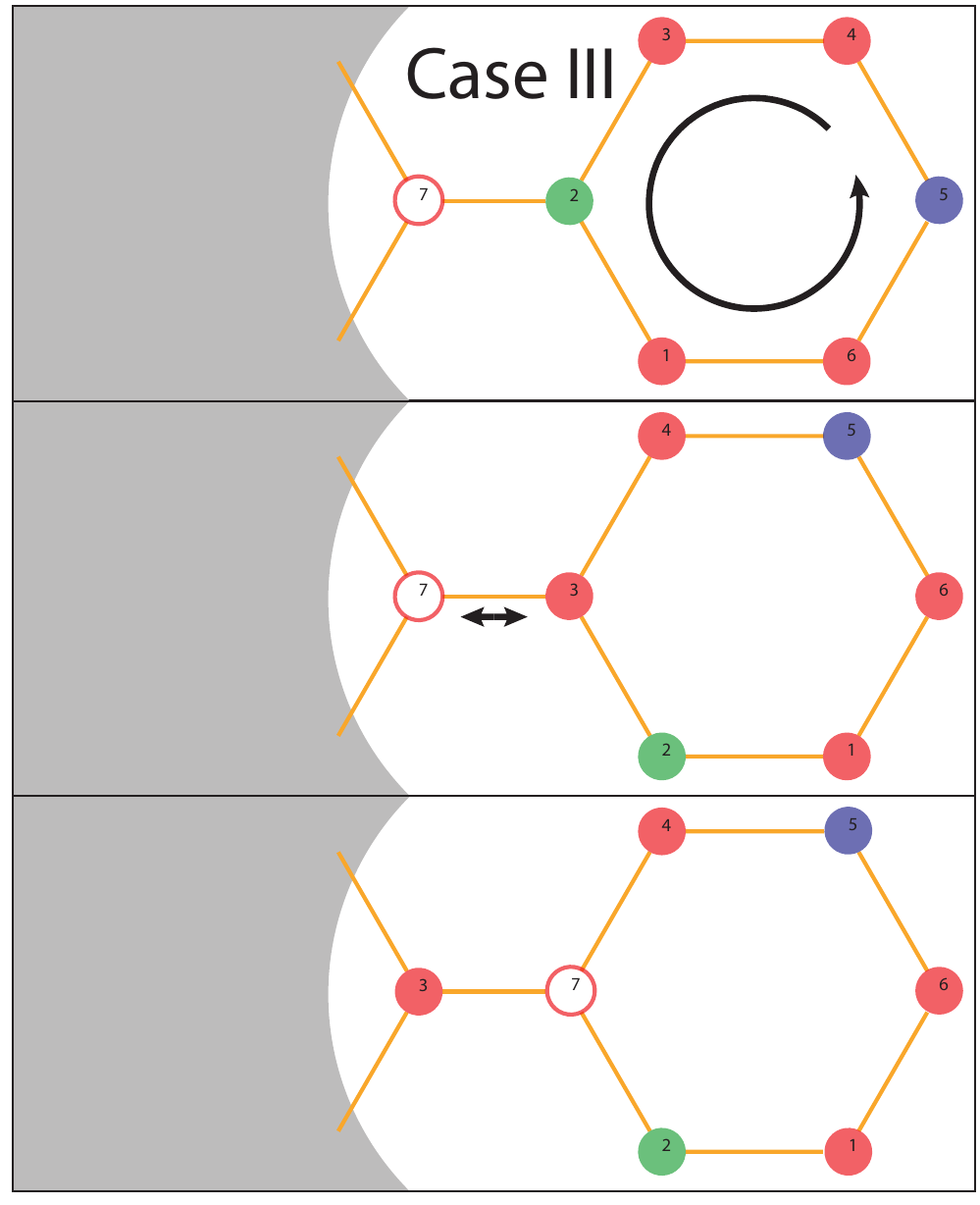}
\vspace{-15px}
\end{wrapfigure}
\TE{Proof of \prettyref{thm:complete}:} Similar to \cite{auletta1999linear}, we can decompose the permutation $\sigma$ into a sequence of $\mathcal{O}(|\VERTEX|)$ pairwise permutations, which has $\mathcal{O}(|\VERTEX|)$ complexity. Further, each pairwise permutation between $\vertex^i$ and $\vertex^j$ can be converted into a sequence of pairwise position swaps between two neighboring vertices along a path connecting $\vertex^i$ and $\vertex^j$, and we assume the shortest path is used. 

\TE{Position Swap:} We construct a sequence of motions to swap the positions of two neighboring vertices $\vertex^i$ and $\vertex^j$, while keeping all other vertices unaltered. Note that we have at least one vacant vertex. We consider two cases based on the loop to which $\vertex^i$ and $\vertex^j$ belong. \TE{Case I:} Suppose that the two vertices belong to two different loops $\vertex^i\in\VERTEXL^k, \vertex^j\in\VERTEXL^l$ ($k\neq l$) and $\vacant\in\VERTEXL^k$. Then we can use the motion sequence in \prettyref{fig:CaseI_II} to accomplish the position swap. \TE{Case II:} If the two vertices as well as $\vacant$ belong to the same loop $\vertex^i,\vertex^j,\vacant\in\VERTEXL^k$, then we can use the motion sequence in \prettyref{fig:CaseI_II} and reduce to \TE{Case I}. \TE{Case III:} If the vacant vertex does not belong to any of the loops of $\vertex^i$ and $\vertex^j$, then we can use \prettyref{lem:vacantMove}, the motion sequence in the inset and reduce to \TE{Case I,II}. Note that each loop has at least 3 vertices so that we can always move $\vacant$ into a loop without affecting the two other vertices to be swapped.

Note that the above motion sequence will alter the positions of irrelevant robots. However, all possible motion sequences will end up in \TE{Case I}, which in turn uses the subsequence in \prettyref{fig:CaseI_II}, and this subsequence does not alter the positions of irrelevant robots. Therefore, we could always reverse the the first half of the motion subsequence afterwards to recover the positions of irrelevant robots. 

\TE{Amortized Motion Sequence Length:} To upper bound the length of motion sequence, we consider the permutation of two (possible non-connected) vertices $\vertex^i$ and $\vertex^j$. The shortest path between $\vertex^i$ and $\vertex^j$ would pass through $K'\leq K$ loops and we denote these loops as:
\begin{align*}
\VERTEXL^1,\cdots,\VERTEXL^{K'},
\end{align*}
where $\vertex^i\in\VERTEXL^1$ and $\vertex^j\in\VERTEXL^{K'}$. We could then pick $K'-2$ intermedite vertices such that:
\begin{align*}
\vertex^m\in\VERTEXL^{m+1}\land m=1,\cdots K'-2.
\end{align*}
The permutation of two vertices $\vertex^i$ and $\vertex^j$ can be replaced by a series of $2K'-3$ permutations: $\vertex^i\leftrightarrow\vertex^1,\cdots,\vertex^i\leftrightarrow\vertex^{K'-2},\vertex^i\leftrightarrow\vertex^j,\vertex^j\leftrightarrow\vertex^{K'-2},\cdots,\vertex^j\leftrightarrow\vertex^1$. Each of these permutations can be accomplished using a similar motion sequence as in \prettyref{fig:CaseI_II}, whose length is upper bounded by the size of two loops. Therefore, for the $2K'-3$ permutations, the total complexity is $\mathcal{O}(4\sum_{k=1}^{K'}\VERTEXL^k)=\mathcal{O}(|\VERTEX|)$, so the length of entire motion sequence is $\mathcal{O}(|\VERTEX|^2)$. 
\begin{algorithm*}[t]
\caption{\revised{MPP}}
\label{alg:planning}
\begin{algorithmic}[1]
\Require{The graph $\mathcal{G}$, initial vertices $\E{s}^{1,\cdots,N}$, goal vertices $\E{t}^{1,\cdots,N}$}
\LineComment{We denote $\E{v}_k^i\leftrightarrow\E{v}_k^j$ as a pebble motion swapping $\E{v}^i$ and $\E{v}^j$ at the $k$th timestep}

\State Initialize a set of swaps $\mathbb{S}\gets\emptyset$\Comment{Stage I: Construct a set of pairwise swaps along graph edges}\label{ln:SwapA}
\For{$i=1,\cdots,N$}
\State Find shortest path $\E{s}^i,\E{v}^1,\cdots,\E{v}^K,\E{t}^i$
\State Decompose the path into swaps $\mathbb{S}\gets\mathbb{S}\bigcup\{\E{s}^i\leftrightarrow\E{v}^1,\cdots,\E{s}^i\leftrightarrow\E{v}^K,\E{s}^i\leftrightarrow\E{t}^i,\E{t}^i\leftrightarrow\E{v}^K,\cdots,\E{t}^i\leftrightarrow\E{v}^1\}$\label{ln:SwapB}
\EndFor

\State Initialize number of timesteps $k\gets0$
\label{ln:PebbleA}
\For{each $\E{v}^i\leftrightarrow\E{v}^j\in\mathbb{S}$}\Comment{Stage II: Implement each swap via pebble motions}
\State Initialize $k\gets k+1$ and $\mathbb{T}_k\gets\emptyset$
\If{$\E{v}^i,\E{v}^j$ belongs to different loop}
\State $\mathbb{T}_k\gets\mathbb{T}_k+\text{CaseI}(\E{v}^i,\E{v}^j)$\Comment{Construct the series of pebble motions in \prettyref{fig:CaseI_II} (Case I)}
\ElsIf{$\E{v}^i,\E{v}^j,\E{v}_\emptyset$ belongs to a same loop}
\State $\mathbb{T}_k\gets\mathbb{T}_k+\text{CaseII}(\E{v}^i,\E{v}^j)$\Comment{Construct the series of pebble motions in \prettyref{fig:CaseI_II} (Case II)}
\Else
\State Select another vertex $\E{v}^m$ in the same loop as $\E{v}^i$, such that $\E{v}^m\neq\E{v}^i, \E{v}^m\neq\E{v}^j$
\State Find shortest path $\E{v}_\emptyset,\E{v}^1,\cdots,\E{v}^K,\E{v}^m$
\State $\mathbb{T}_k\gets\mathbb{T}_k+<\E{v}_\emptyset\leftrightarrow\E{v}^1,\cdots,\E{v}^1\leftrightarrow\E{v}^m>$
\If{$\E{v}^i,\E{v}^j$ belongs to different loop}
\State $\mathbb{T}_k\gets\mathbb{T}_k+\text{CaseI}(\E{v}^i,\E{v}^j)$\Comment{Construct the series of pebble motions in \prettyref{fig:CaseI_II} (Case I)}
\ElsIf{$\E{v}^i,\E{v}^j,\E{v}_\emptyset$ belongs to a same loop}
\State $\mathbb{T}_k\gets\mathbb{T}_k+\text{CaseII}(\E{v}^i,\E{v}^j)$\Comment{Construct the series of pebble motions in \prettyref{fig:CaseI_II} (Case II)}
\EndIf
\EndIf
\LineComment{Reverse all the pebble motions (except for the final swap $\E{v}^i\leftrightarrow\E{v}^j$) to undo affects on other robots}
\State $\mathbb{T}_k\gets\mathbb{T}_k+\text{Reverse}(\mathbb{T}_k-<\E{v}^i\leftrightarrow\E{v}^j>)$
\label{ln:PebbleB}
\EndFor

\For{$t=1,\cdots,k$}\Comment{Stage III: Execute parallel swaps with disjoint vertex set}
\State Initialize parallel swap buffer $\mathbb{B}\gets\emptyset$
\For{$t'=t+1,\cdots,k$}
\If{$\text{VertexSet}(\mathbb{T}_t)\bigcap\text{VertexSet}(\mathbb{T}_{t+1})=\emptyset$}
\State$\mathbb{B}\gets\mathbb{B}\bigcup\{\mathbb{T}_{t'}\}$, remove $\mathbb{T}_{t'}$, $k\gets k-1$
\EndIf
\EndFor
\State ParallelExecute($\mathbb{B}$) and set $\mathbb{B}\gets\emptyset$
\EndFor
\end{algorithmic}
\end{algorithm*}
\begin{algorithm*}[t]
\caption{\revised{ParallelExecute($\mathbb{B}$)}}
\label{alg:parallelExecute}
\begin{algorithmic}[1]
\While{$\mathbb{B}$ is not empty}
\LineComment{We use $\mathbb{B}$ to denote parallel buffer that stores parallel swaps (each swap consists of multiple pebble motions)}
\LineComment{We use $\mathbb{P}$ to denote parallel buffer that stores parallel pebble motions}
\State Initialize pebble motion set $\mathbb{P}\gets\emptyset$
\For{nonempty $\mathbb{T}\in\mathbb{B}$}
\State Remove the first pebble motion $\E{v}^i\leftrightarrow\E{v}^j$ from $\mathbb{T}$, $\mathbb{P}\gets\mathbb{P}\bigcup\{\E{v}^i\leftrightarrow\E{v}^j\}$
\EndFor
\For{$\E{v}^i\leftrightarrow\E{v}^j\in\mathbb{P}$}\Comment{Convert pebble motions to geometric paths}
\If{$\E{v}^i\leftrightarrow\E{v}^j$ is a vacant motion between two loops of an inscribed circle}
\State Find minimal rotational angle to align capsule region and generate trajectories according to \prettyref{fig:typeIISingle}
\ElsIf{Piecewise linear rotation as in \prettyref{fig:shortLoop} is collision-free}
\Comment{$\E{v}^i\leftrightarrow\E{v}^j$ is a rotational motion}
\State Output piecewise linear rotational motion
\Else
\State Output circular rotational motion
\EndIf
\EndFor
\EndWhile
\end{algorithmic}
\end{algorithm*}
\section{\label{sec:planning}Appendix: Planning Algorithm}
\revised{The proof in \prettyref{sec:proof} is constructive, from which we can construct a planning algorithm that answer MPP queries as outlined in \prettyref{alg:planning}, \ref{alg:parallelExecute}. Our algorithm first connect each robot's start and goal positions using a shortest path, and then reduces the robot position permutation to a series of pairwise permutations along graph edges (\prettyref{ln:SwapA} to \prettyref{ln:SwapB}). We then search for a vacant buffer location to assist the swap. If the vacant location does not belong to a neighboring loop, we need to swap it to a neighboring loop by vacant motions. Next, we use the three cases in \prettyref{sec:proof} to further divide the swap into a series of vacant or rotational motions. Finally, since a swap could induces many motions that affect other robots' positions, we reverse the motion sequence to undo these affects and move robots back to their original positions (\prettyref{ln:PebbleA} to \prettyref{ln:PebbleB}).} 

\begin{figure}[ht]
\centering
\scalebox{0.8}{
\includegraphics[width=0.95\linewidth]{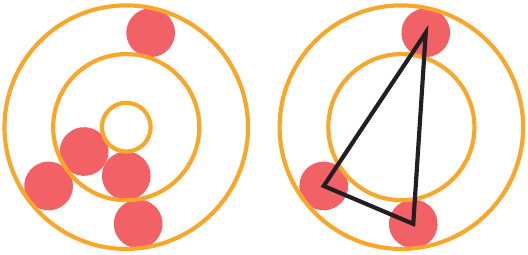}
\put(-130,10){(a)}
\put(-10 ,10){(b)}}
\caption{\small{\label{fig:shortLoop} \revised{Circular paths are necessary in high density (a), while linear path (black) suffice in case of a lower density (b).}}}
\vspace{-5px}
\end{figure}
\revised{The above procedure results in a series of topological pebble motions, and we need to convert each motion into a geometric path for the involved robots. To this end, we use the procedure described in \prettyref{sec:conversion}. However, this would result in redundant motions to rotate and align robots belonging to different loops. We propose two optimization to generate shorter trajectories and a small makespan. First, for two loops that belong to the same inscribed circle, we always find the minimal rotational angle to align the capsule in \prettyref{sec:conversion} to perform vacant moves between loops. Second, our rotational motions follow a circular arc, which ensures collision-free motion when robot density is high. But this is unnecessary with a lower density. In practice, we would compute a shorter piecewise linear loop connecting robot positions of each loop, as illustrated in \prettyref{fig:shortLoop}. We always use the piecewise linear loop whenever it is collision-free. These strategies are summarized in \prettyref{alg:parallelExecute}.}
\fi
\AtNextBibliography{\small}
\printbibliography
\end{document}